\documentclass[11pt,a4paper]{article}
\usepackage[hyperref]{acl2017}
\usepackage{times}
\usepackage{latexsym}

\usepackage{array}
\usepackage{colortbl}

\usepackage{todonotes} 
\newcommand{\barchart}[2]{\begin{tikzpicture}[scale=0.02]
\fill[green] (0,0) rectangle (55,#1);
\fill[blue] (60,0) rectangle (115,#2);
\draw (27.5,0) node[anchor=north] {#1};
\draw (87.5,0) node[anchor=north] {#2};
\end{tikzpicture}}
\newcommand{\rowlabel}[1]{\begin{tikzpicture}[scale=0.02]\fill[white] (0,0) rectangle (2,2);\draw (1,0) node[anchor=north] {\bf #1}; \end{tikzpicture}}

\usepackage{url}

\usepackage{pgfplots}
\pgfplotsset{width=10cm,compat=1.9} 
\definecolor{darkblue}{rgb}{0,0,.5}
\definecolor{darkgreen}{rgb}{0,.5,0}
\definecolor{lightgray}{rgb}{.8,.8,.8}

\aclfinalcopy 

\title{Six Challenges for Neural Machine Translation
}

\author{Philipp Koehn \\
  Computer Science Department\\
  Johns Hopkins University\\
  {\tt phi@jhu.edu} \\\And
  Rebecca Knowles \\
  Computer Science Department\\
  Johns Hopkins University\\
  {\tt rknowles@jhu.edu} \\}

\begin{document}
\maketitle
\begin{abstract}
We explore six challenges for neural machine translation: domain mismatch, amount of training data, rare words, long sentences, word alignment, and beam search. We show both deficiencies and improvements over the quality of phrase-based statistical machine translation.
\end{abstract}

\section{Introduction}

Neural machine translation has emerged as the most promising machine translation approach in recent years, showing superior performance on public benchmarks \citep{bojar-EtAl:2016:WMT1} and rapid adoption in deployments by, e.g., Google \citep{DBLP:journals/corr/WuSCLNMKCGMKSJL16}, Systran \citep{DBLP:journals/corr/CregoKKRYSABCDE16}, and WIPO \citep{IWSLT-2016-Junczys-Dowmunt}. But there have also been reports of poor performance, such as the systems built under low-resource conditions in the DARPA LORELEI program.\footnote{\tt https://www.nist.gov/itl/iad/mig/lorehlt16- evaluations}

In this paper, we examine a number of challenges to neural machine translation (NMT) and give empirical results on how well the technology currently holds up, compared to traditional statistical machine translation (SMT).

We find that:
\begin{enumerate}
\item NMT systems have lower quality {\bf out of domain}, to the point that they completely sacrifice adequacy for the sake of fluency.
\item NMT systems have a steeper learning curve with respect to the {\bf amount of training data}, resulting in worse quality in low-resource settings, but better performance in high-resource settings.
\item NMT systems that operate at the sub-word level (e.g. with byte-pair encoding) perform better than SMT systems on extremely {\bf low-frequency words}, but still show weakness in translating low-frequency words belonging to highly-inflected categories (e.g. verbs).
\item NMT systems have lower translation quality on very {\bf long sentences}, but do comparably better up to a sentence length of about 60 words.
\item The attention model for NMT does not always fulfill the role of a {\bf word alignment model}, but may in fact dramatically diverge.
\item {\bf Beam search decoding} only improves translation quality for narrow beams and deteriorates when exposed to a larger search space.
\end{enumerate}

We note a 7th challenge that we do not examine empirically: NMT systems are much less interpretable. The answer to the question of why the training data leads these systems to decide on specific word choices during decoding is buried in large matrices of real-numbered values. There is a clear need to develop better analytics for NMT.

Other studies have looked at the comparable performance of NMT and SMT systems. \citet{bentivogli-EtAl:2016:EMNLP2016} considered different linguistic categories for English--German and \citet{toral-sanchezcartagena:2017:EACLlong} compared different broad aspects such as fluency and reordering for nine language directions.

\section{Experimental Setup}
We use common toolkits for neural machine translation (Nematus) and traditional phrase-based statistical machine translation (Moses) with common data sets, drawn from WMT and OPUS.

\subsection{Neural Machine Translation}
While a variety of neural machine translation approaches were initially proposed --- such as the use of convolutional neural networks \citep{kalchbrenner-blunsom:2013:EMNLP} --- practically all recent work has been focused on the attention-based encoder-decoder model \citep{bahdanau:ICLR:2015}. 

We use the toolkit Nematus\footnote{\tt https://github.com/rsennrich/nematus/} \citep{sennrich-EtAl:2017:EACLDemo} which has been shown to give state-of-the-art results \citep{WMT16-UEDIN} at the WMT 2016 evaluation campaign \citep{bojar-EtAl:2016:WMT1}.

Unless noted otherwise, we use default settings, such as beam search and single model decoding. The training data is processed with byte-pair encoding \citep{sennrich-haddow-birch:2016:P16-12-2} into subwords to fit a 50,000 word vocabulary limit.

\subsection{Statistical Machine Translation}
Our machine translation systems are trained using Moses\footnote{\tt http://www.stat.org/moses/} \citep{koehn-EtAl:2007:PosterDemo}. We build phrase-based systems using standard features that are commonly used in recent system submissions to WMT \citep{williams-EtAl:2016:WMT,ding-EtAl:2016:WMT}. 

While we use the shorthand SMT for these phrase-based systems, we note that there are other statistical machine translation approaches such as hierarchical phrase-based models \citep{Chiang:CL:2007} and syntax-based models \citep{Galley:2004,galley-EtAl:2006:COLACL} that have been shown to give superior performance for language pairs such as Chinese--English and German--English.

\subsection{Data Conditions}
We carry out our experiments on English--Spanish and German--English. For these language pairs, large training data sets are available.  We use datasets from the shared translation task organized alongside the Conference on Machine Translation (WMT)\footnote{\tt http://www.statmt.org/wmt17/}. For the domain experiments, we use the OPUS corpus\footnote{\tt http://opus.lingfil.uu.se/} \citep{TIEDEMANN12.463}.

Except for the domain experiments, we use the WMT test sets composed of news stories, which are characterized by a broad range of topic, formal language, relatively long sentences (about 30 words on average), and high standards for grammar, orthography, and style.

\section{Challenges}
\subsection{Domain Mismatch}
A known challenge in translation is that in different domains,\footnote{We use the customary definition of domain in machine translation: a {\em domain} is defined by a corpus from a specific source, and may differ from other {\em domains} in topic, genre, style, level of formality, etc.} words have different translations and meaning is expressed in different styles. Hence, a crucial step in developing machine translation systems targeted at a specific use case is domain adaptation. We expect that methods for domain adaptation will be developed for NMT. A currently popular approach is to train a general domain system, followed by training on in-domain data for a few epochs \citep{luong2015stanford,freitag2016fast}.

Often, large amounts of training data are only available out of domain, but we still seek to have robust performance. To test how well NMT and SMT hold up, we trained five different systems using different corpora obtained from OPUS \citep{TIEDEMANN12.463}. An additional system was trained on all the training data. Statistics about corpus sizes are shown in Table~\ref{tab:domain-corpora}. Note that these domains are quite distant from each other, much more so than, say, Europarl, TED Talks, News Commentary, and Global Voices.

\begin{table}
\small
\begin{center}
\begin{tabular}{lrrc}
\bf Corpus & \bf Words & \bf Sentences & \bf W/S \\ \hline
Law (Acquis) &   18,128,173 &  715,372 & 25.3 \\
Medical (EMEA) &  14,301,472 & 1,104,752 & 12.9 \\
IT &   3,041,677 &  337,817 & 9.0 \\
Koran (Tanzil) &   9,848,539 & 480,421 & 20.5 \\
Subtitles & 114,371,754 & 13,873,398 & 8.2 \\
\end{tabular}
\end{center}
\caption{Corpora used to train domain-specific systems, taken from the OPUS repository. IT corpora are GNOME, KDE, PHP, Ubuntu, and OpenOffice.}
\label{tab:domain-corpora}
\end{table}

We trained both SMT and NMT systems for all domains.
All systems were trained for German-English, with tuning and test sets sub-sampled from the data (these were not used in training). A common byte-pair encoding is used for all training runs.

\begin{figure*}
\begin{center}
\begin{tabular}{l|c|c|c|c|c}
\bf System $\downarrow$ & \bf Law & \bf Medical& \bf IT& \bf Koran& \bf Subtitles \\ \hline \hline
\rowlabel{All Data} & \barchart{30.5}{32.8} & \barchart{45.1}{42.2} & \barchart{35.3}{44.7} & \barchart{17.9}{17.9} & \barchart{26.4}{20.8}\\\hline \hline
\rowlabel{Law} & \cellcolor{lightgray}\barchart{31.1}{34.4} & \barchart{12.1}{18.2} & \barchart{3.5}{6.9} & \barchart{1.3}{2.2} & \barchart{2.8}{6.0}\\\hline
\rowlabel{Medical} & \barchart{3.9}{10.2} & \cellcolor{lightgray}\barchart{39.4}{43.5} & \barchart{2.0}{8.5} & \barchart{0.6}{2.0} & \barchart{1.4}{5.8}\\\hline
\rowlabel{IT} & \barchart{1.9}{3.7} & \barchart{6.5}{5.3} & \cellcolor{lightgray}\barchart{42.1}{39.8} & \barchart{1.8}{1.6} & \barchart{3.9}{4.7}\\\hline
\rowlabel{Koran} & \barchart{0.4}{1.8} & \barchart{0.0}{2.1} & \barchart{0.0}{2.3} & \cellcolor{lightgray}\barchart{15.9}{18.8} & \barchart{1.0}{5.5}\\\hline
\rowlabel{Subtitles} & \barchart{7.0}{9.9} & \barchart{9.3}{17.8} & \barchart{9.2}{13.6} & \barchart{9.0}{8.4} & \cellcolor{lightgray}\barchart{25.9}{22.1}\\\hline
\end{tabular}

\end{center}
\caption{Quality of systems (BLEU), when trained on one domain (rows) and tested on another domain (columns). Comparably, NMT systems (left bars) show more degraded performance out of domain.}
\label{tab:domain-results}
\end{figure*}

See Figure~\ref{tab:domain-results} for results. 
While the in-domain NMT and SMT systems are similar (NMT is better for IT and Subtitles, SMT is better for Law, Medical, and Koran), the out-of-domain performance for the NMT systems is worse in almost all cases, sometimes dramatically so. For instance the Medical system leads to a BLEU score of 3.9 (NMT) vs. 10.2 (SMT) on the Law test set. 

Figure~\ref{fig:domain-mismatch-examples} displays an example. When translating the sentence {\em Schaue um dich herum.} (reference: {\em Look around you.}) from the Subtitles corpus, we see mostly non-sensical and completely unrelated output from the NMT system. For instance, the translation from the IT system is {\em Switches to paused.} 

Note that the output of the NMT system is often quite fluent (e.g., {\em Take heed of your own souls.}) but completely unrelated to the input, while the SMT output betrays its difficulties with coping with the out-of-domain input by leaving some words untranslated (e.g., {\em Schaue by dich around.}). This is of particular concern when MT is used for information gisting --- the user will be mislead by hallucinated content in the NMT output.

\begin{figure}
\small
\begin{tabular}{l|p{5.8cm}}
Source & Schaue um dich herum.\\ \hline
Ref. & Look around you. \\ \hline \hline
All & NMT: Look around you.\\
& SMT: Look around you.\\\hline
Law & NMT: Sughum gravecorn.\\
& SMT: In order to implement dich Schaue .\\\hline
Medical & NMT: EMEA / MB / 049 / 01-EN-Final Work progamme for 2002\\
& SMT: Schaue by dich around .\\\hline
IT & NMT: Switches to paused.\\
& SMT: To Schaue by itself . \textbackslash t \textbackslash t\\\hline
Koran & NMT: Take heed of your own souls.\\
& SMT: And you see. \\\hline
Subtitles & NMT: Look around you.\\
& SMT: Look around you .
\end{tabular}
\caption{Examples for the translation of a sentence from the Subtitles corpus, when translated with systems trained on different corpora. Performance out-of-domain is dramatically worse for NMT.}
\label{fig:domain-mismatch-examples}
\end{figure}

\subsection{Amount of Training Data}\label{sec:learning-curve}
A well-known property of statistical systems is that increasing amounts of training data lead to better results. In SMT systems, we have previously observed that doubling the amount of training data gives a fixed increase in BLEU scores. This holds true for both parallel and monolingual data \citep{turchi-debie-cristianini:2008:WMT,irvine-callisonburch:2013:WMT}.

\begin{figure}
\small\noindent\begin{tikzpicture}
\begin{semilogxaxis}[
    title={\bf BLEU Scores with Varying Amounts of Training Data},
    xlabel={Corpus Size (English Words)},
    width=8.7cm,
    height=8cm,
    xmin=300000, xmax=600000000,
    ymin=0, ymax=33,
    xtick={1000000,10000000,100000000},
    ytick={0,10,20,30},
    legend pos=south east,
    ymajorgrids=true,
    grid style=dashed,
]
 
\addplot[
    color=darkblue,
    mark=*,
    nodes near coords,
    nodes near coords align={\thealign},
    visualization depends on={value \thisrow{ALIGN} \as \thealign},
    ]
    table[x=X, y=Y]{
        X    Y ALIGN SIZE 
   377411 21.8 south 1024 
   753253 23.4 south  512  
  1506597 24.9 south  256  
  3013217 26.2 south  128  
  6035176 26.9 south   64  
 12053137 27.9 south   32  
 24107029 28.6 south   16  
 48204905 29.2 south    8  
 96433893 29.6 south    4  
192873315 30.1 north    2 
385703602 30.4 north    1
    };
    \addlegendentry{Phrase-Based with Big LM}

\addplot[
    color=blue,
    mark=square,
    nodes near coords,
    nodes near coords align={\thealign},
    visualization depends on={value \thisrow{ALIGN} \as \thealign},
    ]
    table[x=X, y=Y]{
        X    Y ALIGN SIZE 
   377411 16.4 south 1024 
   753253 18.1 south  512  
  1506597 19.6 south  256  
  3013217 21.2 south  128  
  6035176 22.2 south   64  
 12053137 23.5 south   32  
 24107029 24.7 north   16  
 48204905 26.1 north    8  
 96433893 26.9 north    4  
192873315 27.8 north    2 
385703602 28.6 north    1
    };
    \addlegendentry{Phrase-Based}

\addplot[
    color=darkgreen,
    mark=o,
    nodes near coords,
    nodes near coords align={\thealign},
    visualization depends on={value \thisrow{ALIGN} \as \thealign},
    ]
    table[x=X, y=Y]{
        X    Y ALIGN SIZE 
   377411  1.6 south 1024 
   753253  7.2 south  512  
  1506597 11.9 south  256  
  3013217 14.7 east   128  
  6035176 18.2 east    64  
 12053137 22.4 north   32  
 24107029 25.7 south   16  
 48204905 27.4 south    8  
 96433893 29.2 north    4  
192873315 30.3 south    2 
385703602 31.1 south    1
    };
    \addlegendentry{Neural}
\end{semilogxaxis}
\end{tikzpicture}
\caption{BLEU scores for English-Spanish systems trained on 0.4 million to 385.7 million words of parallel data. Quality for NMT starts much lower, outperforms SMT at about 15 million words, and even beats a SMT system with a big 2 billion word in-domain language model under high-resource conditions.}
\label{fig:learning-curve}
\end{figure}
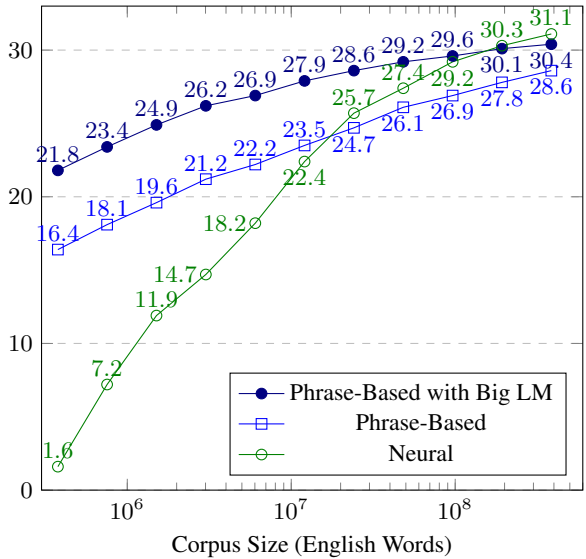

How do the data needs of SMT and NMT compare? NMT promises both to generalize better (exploiting word similary in embeddings) and condition on larger context (entire input and all prior output words).

We built English-Spanish systems on WMT data,\footnote{Spanish was last represented in 2013, we used data from \tt http://statmt.org/wmt13/translation-task.html} about 385.7 million English words paired with Spanish. To obtain a learning curve, we used $\frac{1}{1024}$, $\frac{1}{512}$, ..., $\frac{1}{2}$, and all of the data. For SMT, the language model was trained on the Spanish part of each subset, respectively. In addition to a NMT and SMT system trained on each subset, we also used all additionally provided monolingual data for a big language model in contrastive SMT systems.

Results are shown in Figure~\ref{fig:learning-curve}. NMT exhibits a much steeper learning curve, starting with abysmal results (BLEU score of 1.6 vs. 16.4 for $\frac{1}{1024}$ of the data), outperforming SMT 25.7 vs. 24.7 with $\frac{1}{16}$ of the data (24.1 million words), and even beating the SMT system with a big language model with the full data set (31.1 for NMT, 28.4 for SMT, 30.4 for SMT+BigLM).

The contrast between the NMT and SMT learning curves is quite striking. While NMT is able to exploit increasing amounts of training data more effectively, it is unable to get off the ground with training corpus sizes of a few million words or less. 

To illustrate this, see Figure~\ref{fig:learning-curve-example}. With $\frac{1}{1024}$ of the training data, the output is completely unrelated to the input, some key words are properly translated with $\frac{1}{512}$ and $\frac{1}{256}$ of the data ({\em estrategia} for {\em strategy}, {\em elecci{\'o}n} or {\em elecciones} for {\em election}), and starting with $\frac{1}{64}$ the translations become respectable.

\begin{figure}
\small
\begin{tabular}{l|p{6cm}}
Src: & A Republican strategy to counter the re-election of Obama\\\hline\hline
$\frac{1}{1024}$ & Un {\'o}rgano de coordinaci{\'o}n para el anuncio de libre determinaci{\'o}n\\ \hline
$\frac{1}{512}$ & Lista de una estrategia para luchar contra la elecci{\'o}n de hojas de Ohio\\\hline
$\frac{1}{256}$ & Explosi{\'o}n realiza una estrategia divisiva de luchar contra las elecciones de autor\\\hline
$\frac{1}{128}$ & Una estrategia republicana para la eliminaci{\'o}n de la reelecci{\'o}n de Obama\\\hline
$\frac{1}{64}$ & Estrategia siria para contrarrestar la reelecci{\'o}n del Obama .\\\hline
$\frac{1}{32}+$ & Una estrategia republicana para contrarrestar la reelecci{\'o}n de Obama\\
\end{tabular}
\caption{Translations of the first sentence of the test set using NMT system trained on varying amounts of training data. Under low resource conditions, NMT produces fluent output unrelated to the input.}
\label{fig:learning-curve-example}
\end{figure}

\subsection{Rare Words}
Conventional wisdom states that neural machine translation models perform particularly poorly on rare words, \citep{luong-EtAl:2015:ACL-IJCNLP-2,sennrich-haddow-birch:2016:P16-12-2,discrete} due in part to the smaller vocabularies used by NMT systems.
We examine this claim by comparing performance on rare word translation between NMT and SMT systems of similar quality for German--English and find that NMT systems actually outperform SMT systems on translation of very infrequent words.
However, both NMT and SMT systems do continue to have difficulty translating some infrequent words, particularly those belonging to highly-inflected categories.

\begin{figure*}
\begin{tikzpicture}[scale=11.75]
\small
\fill[darkgreen!60!blue!40!white] (0,0) rectangle (0.01978,0.13222997611);
\fill[darkgreen!80!white] (0,0.13222997611) rectangle (0.01978,0.200737232692);
\fill[darkgreen!60!blue!40!white] (0.01978,0) rectangle (0.02732,0.085529879804);
\fill[darkgreen!80!white] (0.01978,0.085529879804) rectangle (0.02732,0.122895132079);
\fill[darkgreen!60!blue!40!white] (0.02732,0) rectangle (0.03208,0.134848484848);
\fill[darkgreen!80!white] (0.02732,0.134848484848) rectangle (0.03208,0.22012012012);
\fill[darkgreen!60!blue!40!white] (0.03208,0) rectangle (0.03996,0.183976499691);
\fill[darkgreen!80!white] (0.03208,0.183976499691) rectangle (0.03996,0.266620241411);
\fill[darkgreen!60!blue!40!white] (0.03996,0) rectangle (0.04982,0.206659106659);
\fill[darkgreen!80!white] (0.03996,0.206659106659) rectangle (0.04982,0.213786897048);
\fill[darkgreen!60!blue!40!white] (0.04982,0) rectangle (0.06308,0.261277958153);
\fill[blue!30!white] (0.04982,0.261277958153) rectangle (0.06308,0.273362039851);
\fill[darkgreen!60!blue!40!white] (0.06308,0) rectangle (0.08052,0.276048883757);
\fill[darkgreen!80!white] (0.06308,0.276048883757) rectangle (0.08052,0.280376670717);
\fill[darkgreen!60!blue!40!white] (0.08052,0) rectangle (0.10182,0.264346824613);
\fill[blue!30!white] (0.08052,0.264346824613) rectangle (0.10182,0.305581885203);
\fill[darkgreen!60!blue!40!white] (0.10182,0) rectangle (0.13082,0.310074447646);
\fill[blue!30!white] (0.10182,0.310074447646) rectangle (0.13082,0.329311947931);
\fill[darkgreen!60!blue!40!white] (0.13082,0) rectangle (0.16546,0.276802476533);
\fill[blue!30!white] (0.13082,0.276802476533) rectangle (0.16546,0.285416666667);
\fill[darkgreen!60!blue!40!white] (0.16546,0) rectangle (0.20522,0.273818752307);
\fill[blue!30!white] (0.16546,0.273818752307) rectangle (0.20522,0.29184370258);
\fill[darkgreen!60!blue!40!white] (0.20522,0) rectangle (0.25392,0.291074700493);
\fill[blue!30!white] (0.20522,0.291074700493) rectangle (0.25392,0.299135981912);
\fill[darkgreen!60!blue!40!white] (0.25392,0) rectangle (0.31034,0.254877379209);
\fill[blue!30!white] (0.25392,0.254877379209) rectangle (0.31034,0.271422994423);
\fill[darkgreen!60!blue!40!white] (0.31034,0) rectangle (0.37882,0.268617414692);
\fill[blue!30!white] (0.31034,0.268617414692) rectangle (0.37882,0.28606817281);
\fill[darkgreen!60!blue!40!white] (0.37882,0) rectangle (0.4446,0.260455162334);
\fill[blue!30!white] (0.37882,0.260455162334) rectangle (0.4446,0.28561427981);
\fill[darkgreen!60!blue!40!white] (0.4446,0) rectangle (0.52154,0.303029490617);
\fill[blue!30!white] (0.4446,0.303029490617) rectangle (0.52154,0.311076563465);
\fill[darkgreen!60!blue!40!white] (0.52154,0) rectangle (0.58434,0.24728314239);
\fill[darkgreen!80!white] (0.52154,0.24728314239) rectangle (0.58434,0.248513824031);
\fill[darkgreen!60!blue!40!white] (0.58434,0) rectangle (0.6468,0.277647686433);
\fill[blue!30!white] (0.58434,0.277647686433) rectangle (0.6468,0.284259478767);
\fill[darkgreen!60!blue!40!white] (0.6468,0) rectangle (1.2875,0.327062571024);
\fill[darkgreen!80!white] (0.6468,0.327062571024) rectangle (1.2875,0.338964673808);
\draw[darkgreen!70!black,very thick]  (0,0.200737232692) --  (0.01978,0.200737232692) --  (0.01978,0.122895132079) --  (0.02732,0.122895132079) --  (0.02732,0.22012012012) --  (0.03208,0.22012012012) --  (0.03208,0.266620241411) --  (0.03996,0.266620241411) --  (0.03996,0.213786897048) --  (0.04982,0.213786897048) --  (0.04982,0.261277958153) --  (0.06308,0.261277958153) --  (0.06308,0.280376670717) --  (0.08052,0.280376670717) --  (0.08052,0.264346824613) --  (0.10182,0.264346824613) --  (0.10182,0.310074447646) --  (0.13082,0.310074447646) --  (0.13082,0.276802476533) --  (0.16546,0.276802476533) --  (0.16546,0.273818752307) --  (0.20522,0.273818752307) --  (0.20522,0.291074700493) --  (0.25392,0.291074700493) --  (0.25392,0.254877379209) --  (0.31034,0.254877379209) --  (0.31034,0.268617414692) --  (0.37882,0.268617414692) --  (0.37882,0.260455162334) --  (0.4446,0.260455162334) --  (0.4446,0.303029490617) --  (0.52154,0.303029490617) --  (0.52154,0.248513824031) --  (0.58434,0.248513824031) --  (0.58434,0.277647686433) --  (0.6468,0.277647686433) --  (0.6468,0.338964673808) --  (1.2875,0.338964673808);
\draw[blue,very thick]  (0,0.13222997611) --  (0.01978,0.13222997611) --  (0.01978,0.085529879804) --  (0.02732,0.085529879804) --  (0.02732,0.134848484848) --  (0.03208,0.134848484848) --  (0.03208,0.183976499691) --  (0.03996,0.183976499691) --  (0.03996,0.206659106659) --  (0.04982,0.206659106659) --  (0.04982,0.273362039851) --  (0.06308,0.273362039851) --  (0.06308,0.276048883757) --  (0.08052,0.276048883757) --  (0.08052,0.305581885203) --  (0.10182,0.305581885203) --  (0.10182,0.329311947931) --  (0.13082,0.329311947931) --  (0.13082,0.285416666667) --  (0.16546,0.285416666667) --  (0.16546,0.29184370258) --  (0.20522,0.29184370258) --  (0.20522,0.299135981912) --  (0.25392,0.299135981912) --  (0.25392,0.271422994423) --  (0.31034,0.271422994423) --  (0.31034,0.28606817281) --  (0.37882,0.28606817281) --  (0.37882,0.28561427981) --  (0.4446,0.28561427981) --  (0.4446,0.311076563465) --  (0.52154,0.311076563465) --  (0.52154,0.24728314239) --  (0.58434,0.24728314239) --  (0.58434,0.284259478767) --  (0.6468,0.284259478767) --  (0.6468,0.327062571024) --  (1.2875,0.327062571024);

\node[label=below:\rotatebox{90}{\small 0}] at (0.00989,0) {};
\node[label=below:\rotatebox{90}{\small 1---}] at (0.02355,0) {};
\node[label=below:\rotatebox{90}{\small 2------}] at (0.0297,0) {};
\node[label=below:\rotatebox{90}{\small 4}] at (0.03602,0) {};
\node[label=below:\rotatebox{90}{8---}] at (0.04489,0) {};
\node[label=below:\rotatebox{90}{16------}] at (0.05645,0) {};
\node[label=below:\rotatebox{90}{32}] at (0.0718,0) {};
\node[label=below:\rotatebox{90}{64}] at (0.09117,0) {};
\node[label=below:\rotatebox{90}{128}] at (0.11632,0) {};
\node[label=below:\rotatebox{90}{256}] at (0.14814,0) {};
\node[label=below:\rotatebox{90}{512}] at (0.18534,0) {};
\node[label=below:\rotatebox{90}{999}] at (0.22957,0) {};
\node[label=below:\rotatebox{90}{1999}] at (0.28213,0) {};
\node[label=below:\rotatebox{90}{3999}] at (0.34458,0) {};
\node[label=below:\rotatebox{90}{7999}] at (0.41171,0) {};
\node[label=below:\rotatebox{90}{15999}] at (0.48307,0) {};
\node[label=below:\rotatebox{90}{31999}] at (0.55294,0) {};
\node[label=below:\rotatebox{90}{63999}] at (0.61557,0) {};
\node[label=below:\rotatebox{90}{\small 64000+}] at (0.96715,0) {};

\draw (0,0) node[anchor=east] {40\%};
\draw (0,0.1) node[anchor=east] {50\%};
\draw (0,0.2) node[anchor=east] {60\%};
\draw (0,0.3) node[anchor=east] {70\%};
\draw (0,-0.13) node[anchor=east] {0\%};
\draw (0,-0.18) node[anchor=east] {5\%};

\fill[darkgreen!60!blue!40!white] (0,-0.13) rectangle (0.01978,-0.155278058645096);
\fill[darkgreen!80!white] (0,-0.155278058645096) rectangle (0.01978,-0.188645096056623);
\fill[darkgreen!60!blue!40!white] (0.01978,-0.13) rectangle (0.02732,-0.180397877984085);
\fill[darkgreen!80!white] (0.01978,-0.180397877984085) rectangle (0.02732,-0.220185676392573);
\fill[darkgreen!60!blue!40!white] (0.02732,-0.13) rectangle (0.03208,-0.159411764705882);
\fill[darkgreen!80!white] (0.02732,-0.159411764705882) rectangle (0.03208,-0.197226890756303);
\fill[darkgreen!60!blue!40!white] (0.03208,-0.13) rectangle (0.03996,-0.152842639593909);
\fill[darkgreen!80!white] (0.03208,-0.152842639593909) rectangle (0.03996,-0.218832487309645);
\fill[darkgreen!60!blue!40!white] (0.03996,-0.13) rectangle (0.04982,-0.154340770791075);
\fill[darkgreen!80!white] (0.03996,-0.154340770791075) rectangle (0.04982,-0.190851926977688);
\fill[darkgreen!60!blue!40!white] (0.04982,-0.13) rectangle (0.06308,-0.137541478129713);
\fill[darkgreen!80!white] (0.04982,-0.137541478129713) rectangle (0.06308,-0.200889894419306);
\fill[darkgreen!60!blue!40!white] (0.06308,-0.13) rectangle (0.08052,-0.13802752293578);
\fill[darkgreen!80!white] (0.06308,-0.13802752293578) rectangle (0.08052,-0.186192660550459);
\fill[darkgreen!60!blue!40!white] (0.08052,-0.13) rectangle (0.10182,-0.139389671361502);
\fill[darkgreen!80!white] (0.08052,-0.139389671361502) rectangle (0.10182,-0.178826291079812);
\fill[darkgreen!60!blue!40!white] (0.10182,-0.13) rectangle (0.13082,-0.141034482758621);
\fill[darkgreen!80!white] (0.10182,-0.141034482758621) rectangle (0.13082,-0.172758620689655);
\fill[darkgreen!60!blue!40!white] (0.13082,-0.13) rectangle (0.16546,-0.136928406466513);
\fill[darkgreen!80!white] (0.13082,-0.136928406466513) rectangle (0.16546,-0.166374133949192);
\fill[darkgreen!60!blue!40!white] (0.16546,-0.13) rectangle (0.20522,-0.135533199195171);
\fill[darkgreen!80!white] (0.16546,-0.135533199195171) rectangle (0.20522,-0.156659959758551);
\fill[darkgreen!60!blue!40!white] (0.20522,-0.13) rectangle (0.25392,-0.141088295687885);
\fill[darkgreen!80!white] (0.20522,-0.141088295687885) rectangle (0.25392,-0.158747433264887);
\fill[darkgreen!60!blue!40!white] (0.25392,-0.13) rectangle (0.31034,-0.14630627437079);
\fill[darkgreen!80!white] (0.25392,-0.14630627437079) rectangle (0.31034,-0.161549096065225);
\fill[darkgreen!60!blue!40!white] (0.31034,-0.13) rectangle (0.37882,-0.147523364485981);
\fill[darkgreen!80!white] (0.31034,-0.147523364485981) rectangle (0.37882,-0.164754672897196);
\fill[darkgreen!60!blue!40!white] (0.37882,-0.13) rectangle (0.4446,-0.149154758285193);
\fill[darkgreen!80!white] (0.37882,-0.149154758285193) rectangle (0.4446,-0.159188203101247);
\fill[darkgreen!60!blue!40!white] (0.4446,-0.13) rectangle (0.52154,-0.147676111255524);
\fill[darkgreen!80!white] (0.4446,-0.147676111255524) rectangle (0.52154,-0.160413309072004);
\fill[darkgreen!60!blue!40!white] (0.52154,-0.13) rectangle (0.58434,-0.157070063694268);
\fill[darkgreen!80!white] (0.52154,-0.157070063694268) rectangle (0.58434,-0.175222929936306);
\fill[darkgreen!60!blue!40!white] (0.58434,-0.13) rectangle (0.6468,-0.165222542427153);
\fill[darkgreen!80!white] (0.58434,-0.165222542427153) rectangle (0.6468,-0.179631764329171);
\fill[darkgreen!60!blue!40!white] (0.6468,-0.13) rectangle (1.2875,-0.182505072576869);
\fill[darkgreen!80!white] (0.6468,-0.182505072576869) rectangle (1.2875,-0.20747775870142);

\draw[darkgreen!80!black,very thick]  (0,-0.188645096056623) --  (0.01978,-0.188645096056623) --  (0.01978,-0.220185676392573) --  (0.02732,-0.220185676392573) --  (0.02732,-0.197226890756303) --  (0.03208,-0.197226890756303) --  (0.03208,-0.218832487309645) --  (0.03996,-0.218832487309645) --  (0.03996,-0.190851926977688) --  (0.04982,-0.190851926977688) --  (0.04982,-0.200889894419306) --  (0.06308,-0.200889894419306) --  (0.06308,-0.186192660550459) --  (0.08052,-0.186192660550459) --  (0.08052,-0.178826291079812) --  (0.10182,-0.178826291079812) --  (0.10182,-0.172758620689655) --  (0.13082,-0.172758620689655) --  (0.13082,-0.166374133949192) --  (0.16546,-0.166374133949192) --  (0.16546,-0.156659959758551) --  (0.20522,-0.156659959758551) --  (0.20522,-0.158747433264887) --  (0.25392,-0.158747433264887) --  (0.25392,-0.161549096065225) --  (0.31034,-0.161549096065225) --  (0.31034,-0.164754672897196) --  (0.37882,-0.164754672897196) --  (0.37882,-0.159188203101247) --  (0.4446,-0.159188203101247) --  (0.4446,-0.160413309072004) --  (0.52154,-0.160413309072004) --  (0.52154,-0.175222929936306) --  (0.58434,-0.175222929936306) --  (0.58434,-0.179631764329171) --  (0.6468,-0.179631764329171) --  (0.6468,-0.20747775870142) --  (1.2875,-0.20747775870142);
\draw[blue,very thick]  (0,-0.155278058645096) --  (0.01978,-0.155278058645096) --  (0.01978,-0.180397877984085) --  (0.02732,-0.180397877984085) --  (0.02732,-0.159411764705882) --  (0.03208,-0.159411764705882) --  (0.03208,-0.152842639593909) --  (0.03996,-0.152842639593909) --  (0.03996,-0.154340770791075) --  (0.04982,-0.154340770791075) --  (0.04982,-0.137541478129713) --  (0.06308,-0.137541478129713) --  (0.06308,-0.13802752293578) --  (0.08052,-0.13802752293578) --  (0.08052,-0.139389671361502) --  (0.10182,-0.139389671361502) --  (0.10182,-0.141034482758621) --  (0.13082,-0.141034482758621) --  (0.13082,-0.136928406466513) --  (0.16546,-0.136928406466513) --  (0.16546,-0.135533199195171) --  (0.20522,-0.135533199195171) --  (0.20522,-0.141088295687885) --  (0.25392,-0.141088295687885) --  (0.25392,-0.14630627437079) --  (0.31034,-0.14630627437079) --  (0.31034,-0.147523364485981) --  (0.37882,-0.147523364485981) --  (0.37882,-0.149154758285193) --  (0.4446,-0.149154758285193) --  (0.4446,-0.147676111255524) --  (0.52154,-0.147676111255524) --  (0.52154,-0.157070063694268) --  (0.58434,-0.157070063694268) --  (0.58434,-0.165222542427153) --  (0.6468,-0.165222542427153) --  (0.6468,-0.182505072576869) --  (1.2875,-0.182505072576869);
\end{tikzpicture}
\caption{Precision of translation and deletion rates by source words type. SMT (light blue) and NMT (dark green). The horizontal axis represents the corpus frequency of the source types, with the axis labels showing the upper end of the bin. Bin width is proportional to the number of word types in that frequency range. The upper part of the graph shows the precision averaged across all word types in the bin. The lower part shows the proportion of source tokens in the bin that were deleted.}
\label{fig:cc}
\end{figure*}
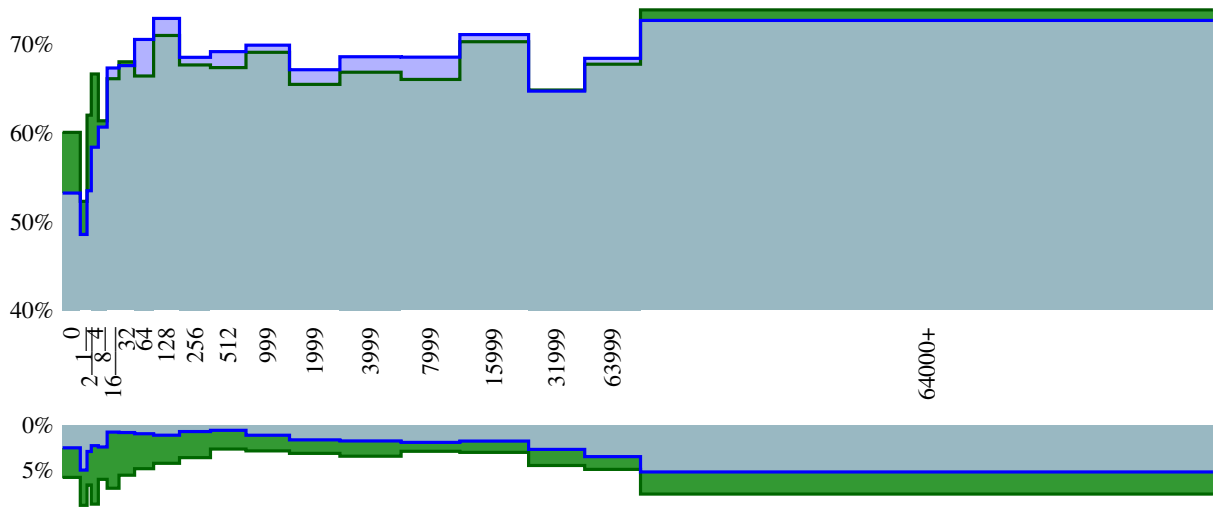

For the neural machine translation model, we use a publicly available model\footnote{\tt https://github.com/rsennrich/wmt16-scripts/} 
with the training settings of Edinburgh's WMT submission \citep{WMT16-UEDIN}. 
This was trained using Nematus\footnote{\tt https://github.com/rsennrich/nematus/} \citep{sennrich-EtAl:2017:EACLDemo}, with byte-pair encodings \citep{sennrich-haddow-birch:2016:P16-12-2} to allow for open-vocabulary NMT.

The phrase-based model that we used was trained using Moses \citep{koehn-EtAl:2007:PosterDemo}, and the training data and parameters match those described in Johns Hopkins University's submission to the WMT shared task \citep{WMT16-JHU}. 

Both models have case-sensitive BLEU scores of 34.5 on the WMT 2016 news test set (for the NMT model, this reflects the BLEU score resulting from translation with a beam size of 1).
We use a single corpus for computing our lexical frequency counts (a concatenation of Common Crawl, Europarl, and News Commentary).

We follow the approach described by \citet{koehn2012interpolated} for examining the effect of source word frequency on translation accuracy.\footnote{First, we automatically align the source sentence and the machine translation output.
We use fast-align \citep{dyer-chahuneau-smith:2013:NAACL-HLT} to align the full training corpus (source and reference) along with the test source and MT output. We use the suggested standard options for alignment and then symmetrize the alignment with grow-diag-final-and.

Each source word is either unaligned (``dropped") or aligned to one or more target language words.
For each target word to which the source word is aligned, we check if that target word appears in the reference translation.
If the target word appears the same number of times in the MT output as in the reference, we award that alignment a score of one.
If the target word appears more times in the MT output than in the reference, we award fractional credit.
If the target word does not appear in the reference, we award zero credit.
We then average these scores over the full set of target words aligned to the given source word to compute the precision for that source word.
Source words can then be binned by frequency and average translation precisions can be computed.}

The overall average precision is quite similar between the NMT and SMT systems, with the SMT system scoring 70.1\% overall and the NMT system scoring 70.3\%.
This reflects the similar overall quality of the MT systems.
Figure \ref{fig:cc} gives a detailed breakdown.
The values above the horizontal axis represent precisions, while the lower portion represents what proportion of the words were deleted.
The first item of note is that the NMT system has an overall higher proportion of deleted words.
Of the 64379 words examined, the NMT system is estimated to have deleted 3769 of them, while the SMT system deleted 2274.
Both the NMT and SMT systems delete very frequent and very infrequent words at higher proportions than words that fall into the middle range.
Across frequencies, the NMT systems delete a higher proportion of words than the SMT system does.
(The related issue of translation length is discussed in more detail in Section \ref{sec:long-sents}.)

The next interesting observation is what happens with unknown words (words which were never observed in the training corpus).
The SMT system translates these correctly 53.2\% of the time, while the NMT system translates them correctly 60.1\% of the time.
This is reflected in Figure \ref{fig:cc}, where the SMT system shows a steep curve up from the unobserved words, while the NMT system does not see a great jump.

\begin{table}
\small
\begin{tabular}{ l | c  c  }
\textbf{Label} & \textbf{Unobserved} & \textbf{Observed Once} \\ \hline
Adjective & 4 & 10 \\
Named Entity & 40 & 42 \\ 
Noun & 35 & 35 \\ 
Number & 12 & 4 \\ 
Verb & 3 & 6 \\ 
Other & 6 & 3 \\
\end{tabular}
\caption{Breakdown of the first 100 tokens that were unobserved in training or observed once in training, by hand-annotated category.}
\label{tab:rare-tags}
\end{table}

Both SMT and NMT systems actually have their worst performance on words that were observed a single time in the training corpus, dropping to 48.6\% and 52.2\%, respectively; even worse than for unobserved words.
Table~\ref{tab:rare-tags} shows a breakdown of the categories of words that were unobserved in the training corpus or observed only once.
The most common categories across both are named entity (including entity and location names) and nouns.
The named entities can often be passed through unchanged (for example, the surname ``Elabdellaoui" is broken into ``E@@ lab@@ d@@ ell@@ a@@ oui" by the byte-pair encoding and is correctly passed through unchanged by both the NMT and SMT systems).
Many of the nouns are compound nouns; when these are correctly translated, it may be attributed to compound-splitting (SMT) or byte-pair encoding (NMT).
The factored SMT system also has access to the stemmed form of words, which can also play a similar role to byte-pair encoding in enabling translation of unobserved inflected forms (e.g. adjectives, verbs).
Unsurprisingly, there are many numbers that were unobserved in the training data; these tend to be translated correctly (with occasional errors due to formatting of commas and periods, resolvable by post-processing).

The categories which involve more extensive inflection (adjectives and verbs) are arguably the most interesting.
Adjectives and verbs have worse accuracy rates and higher deletion rates than nouns across most word frequencies. 
We show examples in Figure~\ref{tab:rare-examples} of situations where the NMT system succeeds and fails, and contrast it with the failures of the SMT system.
In Example 1, the NMT system successfully translates the unobserved adjective \textit{choreographiertes} (choreographed), while the SMT system does not.
In Example 2, the SMT system simply passes the German verb \textit{einkesselte} (closed in on) unchanged into the output, while the NMT system fails silently, selecting the fluent-sounding but semantically inappropriate ``stabbed" instead.

\begin{figure}
\small
\begin{tabular}{l|p{6.1cm}}
Src.&(1) 
... \textbf{choreographiertes} Gesamtkunstwerk ... \\
&(2) ... die Polizei ihn \textbf{einkesselte}. \\\hline
BPE & (1) \textbf{chore@@ ograph@@ iertes} \\ 
 & (2) \textbf{ein@@ kes@@ sel@@ te} \\ \hline
NMT&(1) 
... \textbf{choreographed} overall artwork ...\\
 &(2) ... police \textbf{stabbed} him.\\ \hline
SMT&(1) 
... \textbf{choreographiertes} total work of art ...\\
 &(2) ... police \textbf{einkesselte} him. \\ \hline
Ref.&(1) 
... \textbf{choreographed} complete work of art ...\\
 &(2) ... police \textbf{closed in on} him.
\end{tabular}
\caption{Examples of words that were unobserved in the training corpus, their byte-pair encodings, and their translations.}
\label{tab:rare-examples}
\end{figure}

While there remains room for improvement, NMT systems (at least those using byte-pair encoding) perform better on very low-frequency words then SMT systems do.
Byte-pair encoding is sometimes sufficient (much like stemming or compound-splitting) to allow the successful translation of rare words even though it does not necessarily split words at morphological boundaries.
As with the fluent-sounding but semantically inappropriate examples from domain-mismatch, NMT may sometimes fail similarly when it encounters unknown words even in-domain.

\subsection{Long Sentences}\label{sec:long-sents}
A well-known flaw of early encoder-decoder NMT models was the inability to properly translate long sentences \citep{cho-EtAl:2014:SSST-8,DBLP:journals/corr/Pouget-AbadieBMCB14}. The introduction of the attention model remedied this problem somewhat. But how well?

We used the large English-Spanish system from the learning curve experiments (Section~\ref{sec:learning-curve}), and used it to translate a collection of news test sets from the WMT shared tasks. We broke up these sets into buckets based on source sentence length (1-9 subword tokens, 10-19 subword tokens, etc.) and computed corpus-level BLEU scores for each.

\begin{figure}
\small
\begin{tikzpicture}
\begin{axis}[
    title={\bf BLEU Scores with Varying Sentence Length},
    xlabel={Sentence Length (source, subword count)},
    ylabel={BLEU},
    width=8cm,
    height=6cm,
    xmin=0, xmax=85,
    ymin=25, ymax=37,
    xtick={0,10,20,30,40,50,60,70,80},
    ytick={25,30,35},
    legend pos=south east,
    ymajorgrids=true,
    grid style=dashed,
]
 
\addplot[
    color=darkgreen,
    mark=o,
    nodes near coords,
    nodes near coords align={\thealign},
    visualization depends on={value \thisrow{ALIGN} \as \thealign},
    ]
    table[x=X, y=Y]{
        X    Y ALIGN 
5  27.1 south
15 28.5 south
25 29.6 south
35 31.0 south
45 33.0 south
55 34.7 south
65 34.1 north
75 31.3 north
85 27.7 north
    };
    \addlegendentry{Neural}

\addplot[
    color=blue,
    mark=square,
    nodes near coords,
    nodes near coords align={\thealign},
    visualization depends on={value \thisrow{ALIGN} \as \thealign},
    ]
    table[x=X, y=Y]{
        X    Y ALIGN 
5  26.9 north
15 27.6 north
25 28.7 north
35 30.3 north
45 32.3 north
55 33.8 north
65 34.7 south
75 31.5 south
85 33.9 south
    };
    \addlegendentry{Phrase-Based}

\end{axis}
\end{tikzpicture}
\caption{Quality of translations based on sentence length. SMT outperforms NMT for sentences longer than 60 subword tokens. For very long sentences (80+) quality is much worse due to too short output.}
\label{fig:sentence-length}
\end{figure}
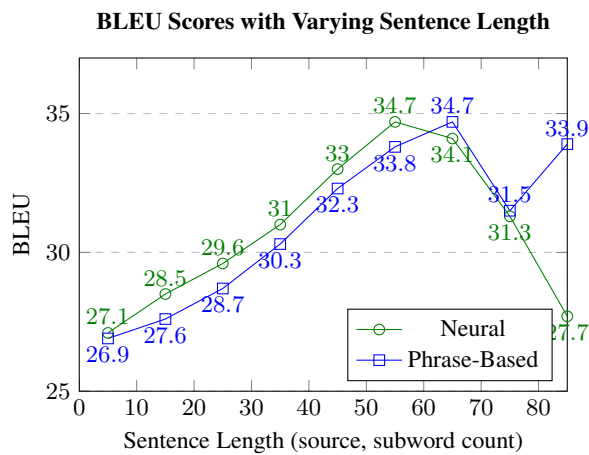

Figure~\ref{fig:sentence-length} shows the results. While overall NMT is better than SMT, the SMT system outperforms NMT on sentences of length 60 and higher. Quality for the two systems is relatively close, except for the very long sentences (80 and more tokens). The quality of the NMT system is dramatically lower for these since it produces too short translations (length ratio 0.859, opposed to 1.024).

\subsection{Word Alignment}
The key contribution of the attention model in neural machine translation \citep{bahdanau:ICLR:2015} was the imposition of an alignment of the output words to the input words. This takes the shape of a probability distribution over the input words which is used to weigh them in a bag-of-words representation of the input sentence.

Arguably, this attention model does not functionally play the role of a word alignment between the source in  the target, at least not in the same way as its analog in statistical machine translation. While in both cases, alignment is a latent variable that is used to obtain probability distributions over words or phrases, arguably the attention model has a broader role. For instance, when translating a verb, attention may also be paid to its subject and object since these may disambiguate it. To further complicate matters, the word representations are products of bidirectional gated recurrent neural networks that have the effect that each word representation is informed by the entire sentence context.

But there is a clear need for an alignment mechanism between source and target words. For instance, prior work used the alignments provided by the attention model to interpolate word translation decisions with traditional probabilistic dictionaries \citep{discrete}, for the introduction of coverage and fertility models \citep{tu-EtAl:2016:P16-1}, etc.

But is the attention model in fact the proper means? To examine this, we compare the soft alignment matrix (the sequence of attention vectors) with word alignments obtained by traditional word alignment methods. We use incremental fast-align \citep{dyer-chahuneau-smith:2013:NAACL-HLT} to align the input and output of the neural machine system. 

\begin{figure}
\small
\begin{tikzpicture}[scale=0.5]
\node[label=above:\rotatebox{90}{relations}] at (0.5,11) {};
\node[label=above:\rotatebox{90}{between}] at (1.5,11) {};
\node[label=above:\rotatebox{90}{Obama}] at (2.5,11) {};
\node[label=above:\rotatebox{90}{and}] at (3.5,11) {};
\node[label=above:\rotatebox{90}{Netanyahu}] at (4.5,11) {};
\node[label=above:\rotatebox{90}{have}] at (5.5,11) {};
\node[label=above:\rotatebox{90}{been}] at (6.5,11) {};
\node[label=above:\rotatebox{90}{strained}] at (7.5,11) {};
\node[label=above:\rotatebox{90}{for}] at (8.5,11) {};
\node[label=above:\rotatebox{90}{years}] at (9.5,11) {};
\node[label=above:\rotatebox{90}{.}] at (10.5,11) {};
\draw (0,10.5) node[anchor=east] {die};
\draw (0,9.5) node[anchor=east] {Beziehungen};
\draw (0,8.5) node[anchor=east] {zwischen};
\draw (0,7.5) node[anchor=east] {Obama};
\draw (0,6.5) node[anchor=east] {und};
\draw (0,5.5) node[anchor=east] {Netanjahu};
\draw (0,4.5) node[anchor=east] {sind};
\draw (0,3.5) node[anchor=east] {seit};
\draw (0,2.5) node[anchor=east] {Jahren};
\draw (0,1.5) node[anchor=east] {angespannt};
\draw (0,0.5) node[anchor=east] {.};
\fill[green!56!white] (0,10) rectangle (1,11);
\draw (0.5,10.5) node[align=center] {56};
\fill[green!89!white] (0,9) rectangle (1,10);
\draw (0.5,9.5) node[align=center] {89};
\fill[green!0!white] (0,8) rectangle (1,9);
\fill[green!0!white] (0,7) rectangle (1,8);
\fill[green!1!white] (0,6) rectangle (1,7);
\fill[green!0!white] (0,5) rectangle (1,6);
\fill[green!0!white] (0,4) rectangle (1,5);
\fill[green!0!white] (0,3) rectangle (1,4);
\fill[green!0!white] (0,2) rectangle (1,3);
\fill[green!0!white] (0,1) rectangle (1,2);
\fill[green!0!white] (0,0) rectangle (1,1);
\fill[green!3!white] (1,10) rectangle (2,11);
\fill[green!1!white] (1,9) rectangle (2,10);
\fill[green!72!white] (1,8) rectangle (2,9);
\draw (1.5,8.5) node[align=center] {72};
\fill[green!2!white] (1,7) rectangle (2,8);
\fill[green!2!white] (1,6) rectangle (2,7);
\fill[green!0!white] (1,5) rectangle (2,6);
\fill[green!0!white] (1,4) rectangle (2,5);
\fill[green!0!white] (1,3) rectangle (2,4);
\fill[green!0!white] (1,2) rectangle (2,3);
\fill[green!0!white] (1,1) rectangle (2,2);
\fill[green!0!white] (1,0) rectangle (2,1);
\fill[green!16!white] (2,10) rectangle (3,11);
\draw (2.5,10.5) node[align=center] {16};
\fill[green!0!white] (2,9) rectangle (3,10);
\fill[green!26!white] (2,8) rectangle (3,9);
\draw (2.5,8.5) node[align=center] {26};
\fill[green!96!white] (2,7) rectangle (3,8);
\draw (2.5,7.5) node[align=center] {96};
\fill[green!1!white] (2,6) rectangle (3,7);
\fill[green!0!white] (2,5) rectangle (3,6);
\fill[green!0!white] (2,4) rectangle (3,5);
\fill[green!0!white] (2,3) rectangle (3,4);
\fill[green!0!white] (2,2) rectangle (3,3);
\fill[green!0!white] (2,1) rectangle (3,2);
\fill[green!0!white] (2,0) rectangle (3,1);
\fill[green!0!white] (3,10) rectangle (4,11);
\fill[green!0!white] (3,9) rectangle (4,10);
\fill[green!0!white] (3,8) rectangle (4,9);
\fill[green!0!white] (3,7) rectangle (4,8);
\fill[green!79!white] (3,6) rectangle (4,7);
\draw (3.5,6.5) node[align=center] {79};
\fill[green!0!white] (3,5) rectangle (4,6);
\fill[green!0!white] (3,4) rectangle (4,5);
\fill[green!0!white] (3,3) rectangle (4,4);
\fill[green!0!white] (3,2) rectangle (4,3);
\fill[green!0!white] (3,1) rectangle (4,2);
\fill[green!0!white] (3,0) rectangle (4,1);
\fill[green!2!white] (4,10) rectangle (5,11);
\fill[green!0!white] (4,9) rectangle (5,10);
\fill[green!0!white] (4,8) rectangle (5,9);
\fill[green!0!white] (4,7) rectangle (5,8);
\fill[green!0!white] (4,6) rectangle (5,7);
\fill[green!98!white] (4,5) rectangle (5,6);
\draw (4.5,5.5) node[align=center] {98};
\fill[green!1!white] (4,4) rectangle (5,5);
\fill[green!2!white] (4,3) rectangle (5,4);
\fill[green!0!white] (4,2) rectangle (5,3);
\fill[green!0!white] (4,1) rectangle (5,2);
\fill[green!0!white] (4,0) rectangle (5,1);
\fill[green!2!white] (5,10) rectangle (6,11);
\fill[green!0!white] (5,9) rectangle (6,10);
\fill[green!0!white] (5,8) rectangle (6,9);
\fill[green!0!white] (5,7) rectangle (6,8);
\fill[green!4!white] (5,6) rectangle (6,7);
\fill[green!0!white] (5,5) rectangle (6,6);
\fill[green!42!white] (5,4) rectangle (6,5);
\draw (5.5,4.5) node[align=center] {42};
\fill[green!3!white] (5,3) rectangle (6,4);
\fill[green!0!white] (5,2) rectangle (6,3);
\fill[green!1!white] (5,1) rectangle (6,2);
\fill[green!11!white] (5,0) rectangle (6,1);
\draw (5.5,0.5) node[align=center] {11};
\fill[green!0!white] (6,10) rectangle (7,11);
\fill[green!0!white] (6,9) rectangle (7,10);
\fill[green!0!white] (6,8) rectangle (7,9);
\fill[green!0!white] (6,7) rectangle (7,8);
\fill[green!2!white] (6,6) rectangle (7,7);
\fill[green!0!white] (6,5) rectangle (7,6);
\fill[green!11!white] (6,4) rectangle (7,5);
\draw (6.5,4.5) node[align=center] {11};
\fill[green!2!white] (6,3) rectangle (7,4);
\fill[green!0!white] (6,2) rectangle (7,3);
\fill[green!4!white] (6,1) rectangle (7,2);
\fill[green!14!white] (6,0) rectangle (7,1);
\draw (6.5,0.5) node[align=center] {14};
\fill[green!6!white] (7,10) rectangle (8,11);
\fill[green!4!white] (7,9) rectangle (8,10);
\fill[green!0!white] (7,8) rectangle (8,9);
\fill[green!0!white] (7,7) rectangle (8,8);
\fill[green!4!white] (7,6) rectangle (8,7);
\fill[green!0!white] (7,5) rectangle (8,6);
\fill[green!38!white] (7,4) rectangle (8,5);
\draw (7.5,4.5) node[align=center] {38};
\fill[green!22!white] (7,3) rectangle (8,4);
\draw (7.5,3.5) node[align=center] {22};
\fill[green!0!white] (7,2) rectangle (8,3);
\fill[green!84!white] (7,1) rectangle (8,2);
\draw (7.5,1.5) node[align=center] {84};
\fill[green!23!white] (7,0) rectangle (8,1);
\draw (7.5,0.5) node[align=center] {23};
\fill[green!8!white] (8,10) rectangle (9,11);
\fill[green!1!white] (8,9) rectangle (9,10);
\fill[green!0!white] (8,8) rectangle (9,9);
\fill[green!0!white] (8,7) rectangle (9,8);
\fill[green!1!white] (8,6) rectangle (9,7);
\fill[green!0!white] (8,5) rectangle (9,6);
\fill[green!1!white] (8,4) rectangle (9,5);
\fill[green!54!white] (8,3) rectangle (9,4);
\draw (8.5,3.5) node[align=center] {54};
\fill[green!0!white] (8,2) rectangle (9,3);
\fill[green!0!white] (8,1) rectangle (9,2);
\fill[green!0!white] (8,0) rectangle (9,1);
\fill[green!1!white] (9,10) rectangle (10,11);
\fill[green!0!white] (9,9) rectangle (10,10);
\fill[green!0!white] (9,8) rectangle (10,9);
\fill[green!0!white] (9,7) rectangle (10,8);
\fill[green!0!white] (9,6) rectangle (10,7);
\fill[green!0!white] (9,5) rectangle (10,6);
\fill[green!0!white] (9,4) rectangle (10,5);
\fill[green!10!white] (9,3) rectangle (10,4);
\draw (9.5,3.5) node[align=center] {10};
\fill[green!98!white] (9,2) rectangle (10,3);
\draw (9.5,2.5) node[align=center] {98};
\fill[green!0!white] (9,1) rectangle (10,2);
\fill[green!0!white] (9,0) rectangle (10,1);
\fill[green!1!white] (10,10) rectangle (11,11);
\fill[green!0!white] (10,9) rectangle (11,10);
\fill[green!0!white] (10,8) rectangle (11,9);
\fill[green!0!white] (10,7) rectangle (11,8);
\fill[green!1!white] (10,6) rectangle (11,7);
\fill[green!0!white] (10,5) rectangle (11,6);
\fill[green!2!white] (10,4) rectangle (11,5);
\fill[green!2!white] (10,3) rectangle (11,4);
\fill[green!0!white] (10,2) rectangle (11,3);
\fill[green!7!white] (10,1) rectangle (11,2);
\fill[green!49!white] (10,0) rectangle (11,1);
\draw (10.5,0.5) node[align=center] {49};
\draw[blue,very thick] (0,9) rectangle (1,10);
\draw[blue,very thick] (1,8) rectangle (2,9);
\draw[blue,very thick] (2,7) rectangle (3,8);
\draw[blue,very thick] (3,6) rectangle (4,7);
\draw[blue,very thick] (4,5) rectangle (5,6);
\draw[blue,very thick] (5,3) rectangle (6,4);
\draw[blue,very thick] (6,4) rectangle (7,5);
\draw[blue,very thick] (6,3) rectangle (7,4);
\draw[blue,very thick] (7,1) rectangle (8,2);
\draw[blue,very thick] (8,3) rectangle (9,4);
\draw[blue,very thick] (9,2) rectangle (10,3);
\draw[blue,very thick] (10,0) rectangle (11,1);


\end{tikzpicture}
\caption{Word alignment for English--German: comparing the attention model states (green boxes with probability in percent if over 10) with alignments obtained from fast-align (blue outlines).}
\label{fig:word-alignment-matrix}
\end{figure}
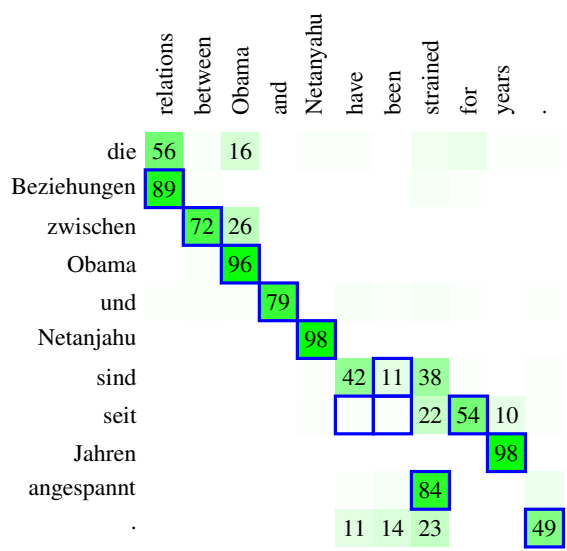

See Figure~\ref{fig:word-alignment-matrix} for an illustration. We compare the word attention states (green boxes) with the word alignments obtained with fast align (blue outlines). For most words, these match up pretty well. Both attention states and fast-align alignment points are a bit fuzzy around the function words {\em have-been/sind}.

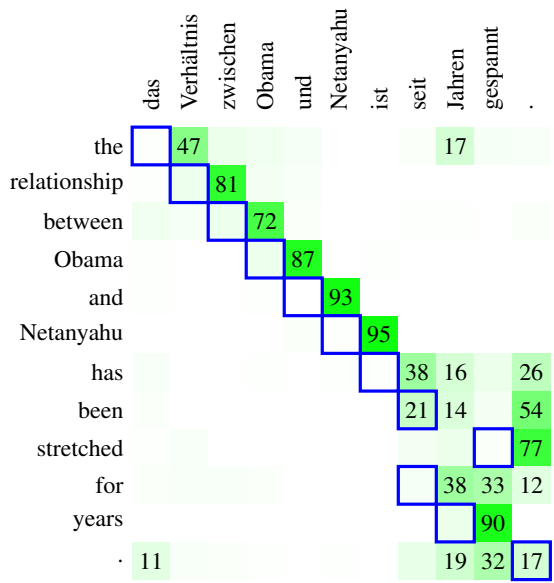
\begin{figure}
\small
\begin{tikzpicture}[scale=0.5]
\node[label=above:\rotatebox{90}{das}] at (0.5,12) {};
\node[label=above:\rotatebox{90}{Verh\"altnis}] at (1.5,12) {};
\node[label=above:\rotatebox{90}{zwischen}] at (2.5,12) {};
\node[label=above:\rotatebox{90}{Obama}] at (3.5,12) {};
\node[label=above:\rotatebox{90}{und}] at (4.5,12) {};
\node[label=above:\rotatebox{90}{Netanyahu}] at (5.5,12) {};
\node[label=above:\rotatebox{90}{ist}] at (6.5,12) {};
\node[label=above:\rotatebox{90}{seit}] at (7.5,12) {};
\node[label=above:\rotatebox{90}{Jahren}] at (8.5,12) {};
\node[label=above:\rotatebox{90}{gespannt}] at (9.5,12) {};
\node[label=above:\rotatebox{90}{.}] at (10.5,12) {};
\draw (0,11.5) node[anchor=east] {the};
\draw (0,10.5) node[anchor=east] {relationship};
\draw (0,9.5) node[anchor=east] {between};
\draw (0,8.5) node[anchor=east] {Obama};
\draw (0,7.5) node[anchor=east] {and};
\draw (0,6.5) node[anchor=east] {Netanyahu};
\draw (0,5.5) node[anchor=east] {has};
\draw (0,4.5) node[anchor=east] {been};
\draw (0,3.5) node[anchor=east] {stretched};
\draw (0,2.5) node[anchor=east] {for};
\draw (0,1.5) node[anchor=east] {years};
\draw (0,0.5) node[anchor=east] {.};
\fill[green!1!white] (0,11) rectangle (1,12);
\fill[green!1!white] (0,10) rectangle (1,11);
\fill[green!6!white] (0,9) rectangle (1,10);
\fill[green!1!white] (0,8) rectangle (1,9);
\fill[green!1!white] (0,7) rectangle (1,8);
\fill[green!0!white] (0,6) rectangle (1,7);
\fill[green!3!white] (0,5) rectangle (1,6);
\fill[green!2!white] (0,4) rectangle (1,5);
\fill[green!1!white] (0,3) rectangle (1,4);
\fill[green!2!white] (0,2) rectangle (1,3);
\fill[green!0!white] (0,1) rectangle (1,2);
\fill[green!11!white] (0,0) rectangle (1,1);
\draw (0.5,0.5) node[align=center] {11};
\fill[green!47!white] (1,11) rectangle (2,12);
\draw (1.5,11.5) node[align=center] {47};
\fill[green!7!white] (1,10) rectangle (2,11);
\fill[green!4!white] (1,9) rectangle (2,10);
\fill[green!0!white] (1,8) rectangle (2,9);
\fill[green!0!white] (1,7) rectangle (2,8);
\fill[green!0!white] (1,6) rectangle (2,7);
\fill[green!0!white] (1,5) rectangle (2,6);
\fill[green!0!white] (1,4) rectangle (2,5);
\fill[green!3!white] (1,3) rectangle (2,4);
\fill[green!3!white] (1,2) rectangle (2,3);
\fill[green!0!white] (1,1) rectangle (2,2);
\fill[green!3!white] (1,0) rectangle (2,1);
\fill[green!8!white] (2,11) rectangle (3,12);
\fill[green!81!white] (2,10) rectangle (3,11);
\draw (2.5,10.5) node[align=center] {81};
\fill[green!7!white] (2,9) rectangle (3,10);
\fill[green!0!white] (2,8) rectangle (3,9);
\fill[green!0!white] (2,7) rectangle (3,8);
\fill[green!0!white] (2,6) rectangle (3,7);
\fill[green!0!white] (2,5) rectangle (3,6);
\fill[green!0!white] (2,4) rectangle (3,5);
\fill[green!0!white] (2,3) rectangle (3,4);
\fill[green!3!white] (2,2) rectangle (3,3);
\fill[green!0!white] (2,1) rectangle (3,2);
\fill[green!2!white] (2,0) rectangle (3,1);
\fill[green!6!white] (3,11) rectangle (4,12);
\fill[green!5!white] (3,10) rectangle (4,11);
\fill[green!72!white] (3,9) rectangle (4,10);
\draw (3.5,9.5) node[align=center] {72};
\fill[green!7!white] (3,8) rectangle (4,9);
\fill[green!1!white] (3,7) rectangle (4,8);
\fill[green!0!white] (3,6) rectangle (4,7);
\fill[green!1!white] (3,5) rectangle (4,6);
\fill[green!0!white] (3,4) rectangle (4,5);
\fill[green!0!white] (3,3) rectangle (4,4);
\fill[green!2!white] (3,2) rectangle (4,3);
\fill[green!0!white] (3,1) rectangle (4,2);
\fill[green!1!white] (3,0) rectangle (4,1);
\fill[green!4!white] (4,11) rectangle (5,12);
\fill[green!3!white] (4,10) rectangle (5,11);
\fill[green!2!white] (4,9) rectangle (5,10);
\fill[green!87!white] (4,8) rectangle (5,9);
\draw (4.5,8.5) node[align=center] {87};
\fill[green!0!white] (4,7) rectangle (5,8);
\fill[green!2!white] (4,6) rectangle (5,7);
\fill[green!0!white] (4,5) rectangle (5,6);
\fill[green!0!white] (4,4) rectangle (5,5);
\fill[green!0!white] (4,3) rectangle (5,4);
\fill[green!0!white] (4,2) rectangle (5,3);
\fill[green!0!white] (4,1) rectangle (5,2);
\fill[green!0!white] (4,0) rectangle (5,1);
\fill[green!0!white] (5,11) rectangle (6,12);
\fill[green!0!white] (5,10) rectangle (6,11);
\fill[green!0!white] (5,9) rectangle (6,10);
\fill[green!0!white] (5,8) rectangle (6,9);
\fill[green!93!white] (5,7) rectangle (6,8);
\draw (5.5,7.5) node[align=center] {93};
\fill[green!1!white] (5,6) rectangle (6,7);
\fill[green!1!white] (5,5) rectangle (6,6);
\fill[green!0!white] (5,4) rectangle (6,5);
\fill[green!0!white] (5,3) rectangle (6,4);
\fill[green!0!white] (5,2) rectangle (6,3);
\fill[green!0!white] (5,1) rectangle (6,2);
\fill[green!1!white] (5,0) rectangle (6,1);
\fill[green!0!white] (6,11) rectangle (7,12);
\fill[green!0!white] (6,10) rectangle (7,11);
\fill[green!0!white] (6,9) rectangle (7,10);
\fill[green!1!white] (6,8) rectangle (7,9);
\fill[green!0!white] (6,7) rectangle (7,8);
\fill[green!95!white] (6,6) rectangle (7,7);
\draw (6.5,6.5) node[align=center] {95};
\fill[green!1!white] (6,5) rectangle (7,6);
\fill[green!0!white] (6,4) rectangle (7,5);
\fill[green!0!white] (6,3) rectangle (7,4);
\fill[green!0!white] (6,2) rectangle (7,3);
\fill[green!0!white] (6,1) rectangle (7,2);
\fill[green!0!white] (6,0) rectangle (7,1);
\fill[green!2!white] (7,11) rectangle (8,12);
\fill[green!0!white] (7,10) rectangle (8,11);
\fill[green!1!white] (7,9) rectangle (8,10);
\fill[green!0!white] (7,8) rectangle (8,9);
\fill[green!0!white] (7,7) rectangle (8,8);
\fill[green!0!white] (7,6) rectangle (8,7);
\fill[green!38!white] (7,5) rectangle (8,6);
\draw (7.5,5.5) node[align=center] {38};
\fill[green!21!white] (7,4) rectangle (8,5);
\draw (7.5,4.5) node[align=center] {21};
\fill[green!5!white] (7,3) rectangle (8,4);
\fill[green!4!white] (7,2) rectangle (8,3);
\fill[green!0!white] (7,1) rectangle (8,2);
\fill[green!9!white] (7,0) rectangle (8,1);
\fill[green!17!white] (8,11) rectangle (9,12);
\draw (8.5,11.5) node[align=center] {17};
\fill[green!0!white] (8,10) rectangle (9,11);
\fill[green!1!white] (8,9) rectangle (9,10);
\fill[green!0!white] (8,8) rectangle (9,9);
\fill[green!0!white] (8,7) rectangle (9,8);
\fill[green!0!white] (8,6) rectangle (9,7);
\fill[green!16!white] (8,5) rectangle (9,6);
\draw (8.5,5.5) node[align=center] {16};
\fill[green!14!white] (8,4) rectangle (9,5);
\draw (8.5,4.5) node[align=center] {14};
\fill[green!8!white] (8,3) rectangle (9,4);
\fill[green!38!white] (8,2) rectangle (9,3);
\draw (8.5,2.5) node[align=center] {38};
\fill[green!8!white] (8,1) rectangle (9,2);
\fill[green!19!white] (8,0) rectangle (9,1);
\draw (8.5,0.5) node[align=center] {19};
\fill[green!4!white] (9,11) rectangle (10,12);
\fill[green!0!white] (9,10) rectangle (10,11);
\fill[green!0!white] (9,9) rectangle (10,10);
\fill[green!0!white] (9,8) rectangle (10,9);
\fill[green!0!white] (9,7) rectangle (10,8);
\fill[green!0!white] (9,6) rectangle (10,7);
\fill[green!8!white] (9,5) rectangle (10,6);
\fill[green!5!white] (9,4) rectangle (10,5);
\fill[green!2!white] (9,3) rectangle (10,4);
\fill[green!33!white] (9,2) rectangle (10,3);
\draw (9.5,2.5) node[align=center] {33};
\fill[green!90!white] (9,1) rectangle (10,2);
\draw (9.5,1.5) node[align=center] {90};
\fill[green!32!white] (9,0) rectangle (10,1);
\draw (9.5,0.5) node[align=center] {32};
\fill[green!3!white] (10,11) rectangle (11,12);
\fill[green!0!white] (10,10) rectangle (11,11);
\fill[green!2!white] (10,9) rectangle (11,10);
\fill[green!0!white] (10,8) rectangle (11,9);
\fill[green!0!white] (10,7) rectangle (11,8);
\fill[green!0!white] (10,6) rectangle (11,7);
\fill[green!26!white] (10,5) rectangle (11,6);
\draw (10.5,5.5) node[align=center] {26};
\fill[green!54!white] (10,4) rectangle (11,5);
\draw (10.5,4.5) node[align=center] {54};
\fill[green!77!white] (10,3) rectangle (11,4);
\draw (10.5,3.5) node[align=center] {77};
\fill[green!12!white] (10,2) rectangle (11,3);
\draw (10.5,2.5) node[align=center] {12};
\fill[green!0!white] (10,1) rectangle (11,2);
\fill[green!17!white] (10,0) rectangle (11,1);
\draw (10.5,0.5) node[align=center] {17};
\draw[blue,very thick] (0,11) rectangle (1,12);
\draw[blue,very thick] (1,10) rectangle (2,11);
\draw[blue,very thick] (2,9) rectangle (3,10);
\draw[blue,very thick] (3,8) rectangle (4,9);
\draw[blue,very thick] (4,7) rectangle (5,8);
\draw[blue,very thick] (5,6) rectangle (6,7);
\draw[blue,very thick] (6,5) rectangle (7,6);
\draw[blue,very thick] (7,4) rectangle (8,5);
\draw[blue,very thick] (7,2) rectangle (8,3);
\draw[blue,very thick] (8,1) rectangle (9,2);
\draw[blue,very thick] (9,3) rectangle (10,4);
\draw[blue,very thick] (10,0) rectangle (11,1);


\end{tikzpicture}
\caption{Mismatch between attention states and desired word alignments (German--English).}
\label{fig:flaky-word-alignment-matrix}
\end{figure}

However, the attention model may settle on alignments that do not correspond with our intuition or alignment points obtained with fast-align. See Figure~\ref{fig:flaky-word-alignment-matrix} for the reverse language direction, German--English. All the alignment points appear to be off by one position. We are not aware of any intuitive explanation for this divergent behavior --- the translation quality is high for both systems.

We measure how well the soft alignment (attention model) of the NMT system match the alignments of fast-align with two metrics:
\begin{itemize}
\item a {\bf match score} that checks for each output if the aligned input word according to fast-align is indeed the input word that received the highest attention probability, and
\item a {\bf probability mass score} that sums up the probability mass given to each alignment point obtained from fast-align.
\end{itemize}
In these scores, we have to handle byte pair encoding and many-to-many alignments\footnote{(1) NMT operates on subwords, but fast-align is run on full words. (2) If an input word is split into subwords by byte pair encoding, then we add their attention scores. (3) If an output word is split into subwords, then we take the average of their attention vectors. (4) The match scores and probability mass scores are computed as average over output word-level scores. (5) If an output word has no fast-align alignment point, it is ignored in this computation. (6) If an output word is fast-aligned to multiple input words, then (6a) for the match score: count it as correct if the $n$ aligned words among the top $n$ highest scoring words according to attention and (6b) for the probability mass score: add up their attention scores.}

In out experiment, we use the neural machine translation models provided by Edinburgh\footnote{\tt https://github.com/rsennrich/wmt16-scripts} \citep{WMT16-UEDIN}. We run fast-align on the same parallel data sets to obtain alignment models and used them to align the input and output of the NMT system.
Table~\ref{tab:word-alignment-scores} shows alignment scores for the systems. The results suggest that, while drastic, the divergence for German--English is an outlier. We note, however, that we have seen such large a divergence also under different data conditions.

\begin{table}
\begin{center}
\begin{tabular}{l|c|c}
\bf Language Pair & Match & Prob. \\ \hline
German--English & 14.9\% & 16.0\%\\
English--German & 77.2\% & 63.2\% \\
Czech--English & 78.0\% & 63.3\% \\
English--Czech & 76.1\% & 59.7\% \\
Russian--English & 72.5\% & 65.0\% \\
English--Russian & 73.4\% & 64.1\% \\
\end{tabular}
\end{center}
\caption{Scores indicating overlap between attention probabilities and alignments obtained with fast-align.}
\label{tab:word-alignment-scores}
\end{table}

Note that the attention model may produce better word alignments by guided alignment training \citep{DBLP:journals/corr/ChenMKP16,liu-EtAl:2016:COLING} where supervised word alignments (such as the ones produced by fast-align) are provided to model training.

\subsection{Beam Search}
\begin{figure*}
\input{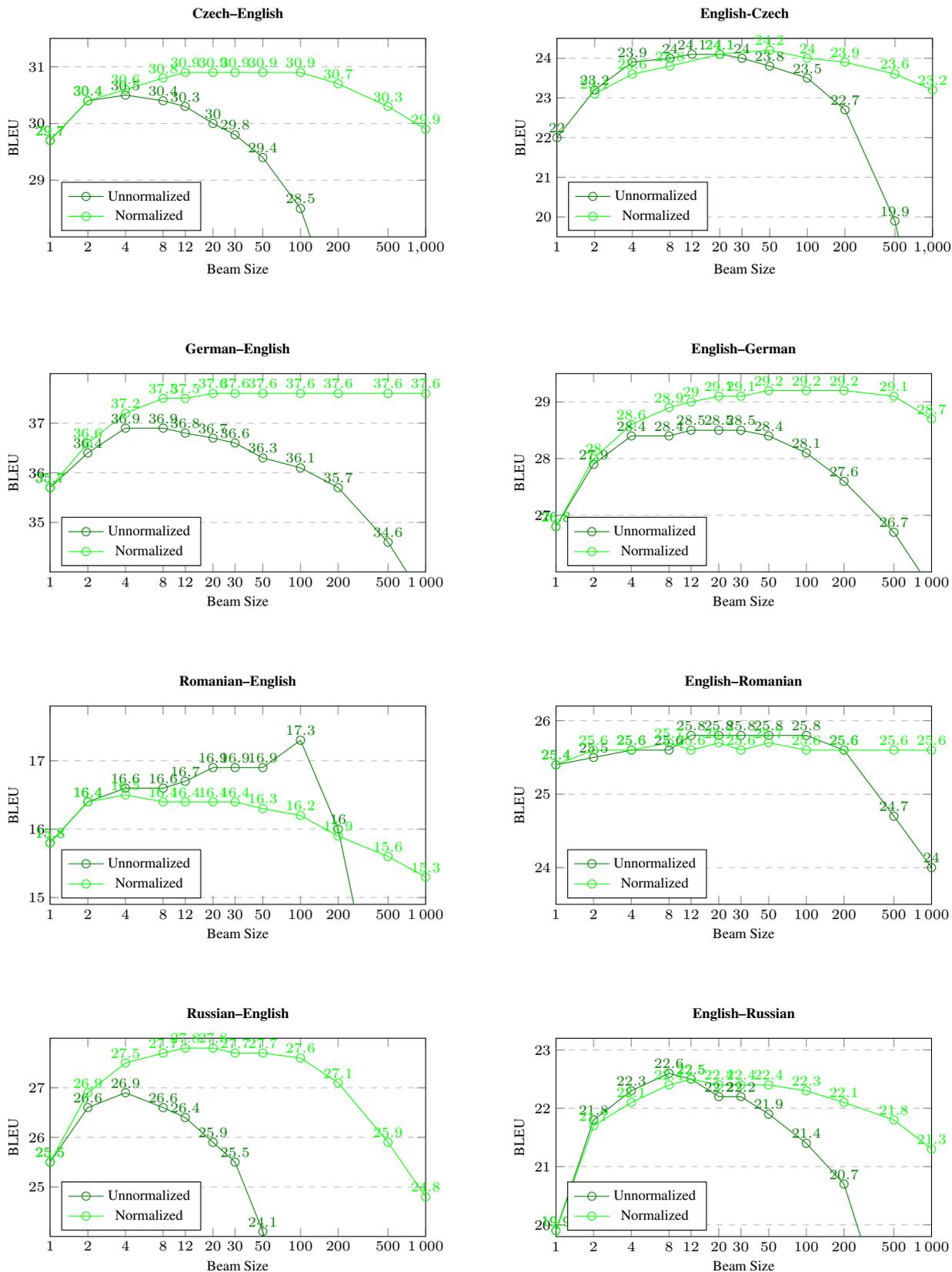}
\caption{Translation quality with varying beam sizes. For large beams, quality decreases, especially when not normalizing scores by sentence length.}
\label{fig:beam-search}
\end{figure*}

The task of decoding is to find the full sentence translation with the highest probability. In statistical machine translation, this problem has been addressed with heuristic search techniques that explore a subset of the space of possible translation. A common feature of these search techniques is a beam size parameter that limits the number of partial translations maintained per input word.

There is typically a straightforward relationship between this beam size parameter and the model score of resulting translations and also their quality score (e.g., BLEU). While there are diminishing returns for increasing the beam parameter, typically improvements in these scores can be expected with larger beams.

Decoding in neural translation models can be set up in similar fashion. When predicting the next output word, we may not only commit to the highest scoring word prediction but also maintain the next best scoring words in a list of partial translations. We record with each partial translation the word translation probabilities (obtained from the softmax), extend each partial translation with subsequent word predictions and accumulate these scores. Since the number of partial translation explodes exponentially with each new output word, we prune them down to a beam of highest scoring partial translations.

As in traditional statistical machine translation decoding, increasing the beam size allows us to explore a larger set of the space of possible translation and hence find translations with better model scores.

However, as Figure~\ref{fig:beam-search} illustrates, increasing the beam size does not consistently improve translation quality. In fact, in almost all cases, worse translations are found beyond an optimal beam size setting (we are using again Edinburgh's WMT 2016 systems). The optimal beam size varies from 4 (e.g., Czech--English) to around 30 (English--Romanian).

Normalizing sentence level model scores by length of the output alleviates the problem somewhat and also leads to better optimal quality in most cases (5 of the 8 language pairs investigated). Optimal beam sizes are in the range of 30--50 in almost all cases, but quality still drops with larger beams. The main cause of deteriorating quality are shorter translations under wider beams.

\section{Conclusions}
We showed that, despite its recent successes, neural machine translation still has to overcome various challenges, most notably performance out-of-domain and under low resource conditions. We hope that this paper motivates research to address these challenges.

What a lot of the problems have in common is that the neural translation models do not show robust behavior when confronted with conditions that differ significantly from training conditions --- may it be due to limited exposure to training data, unusual input in case of out-of-domain test sentences, or unlikely initial word choices in beam search. The solution to these problems may hence lie in a more general approach of training that steps outside optimizing single word predictions given perfectly matching prior sequences.

\section*{Acknowledgment}
This work was partially supported by a Amazon Research Award (to the first author) and a National Science Foundation Graduate Research Fellowship under Grant No. DGE-1232825 (to the second author).

\bibliographystyle{acl_natbib}
\bibliography{mt,more,acl2017}
\end{document}